\documentclass[lettersize,journal]{IEEEtran}

\usepackage{amsmath,amsfonts}
\usepackage{algorithmic}
\usepackage{algorithm}
\usepackage{array}
\usepackage[caption=false,font=normalsize,labelfont=sf,textfont=sf]{subfig}
\usepackage{textcomp}
\usepackage{stfloats}
\usepackage{url}
\usepackage{verbatim}
\usepackage{graphicx}
\usepackage{tabularx}
\usepackage{soul}
\usepackage{multirow}
\usepackage{arydshln}
\usepackage{makecell}
\usepackage{contour}
\usepackage{amssymb}
\usepackage{color}
\usepackage{colortbl}
\usepackage{xcolor}
\usepackage{tabularx}
\usepackage{booktabs}
\usepackage{enumitem}
\definecolor{darkgreen}{RGB}{208, 225, 209}
\definecolor{CA}{RGB}{187, 108, 83}
\definecolor{LR}{RGB}{82, 110, 116}
\definecolor{STANCE}{RGB}{89, 153, 141}
\usepackage{pifont}

\begin{document}

\title{MT$^2$-CSD: A New Dataset and Multi-Semantic Knowledge Fusion Method for Conversational Stance Detection}

\author{
        Fuqiang~Niu$^\dag$,
        Genan~Dai$^\dag$,
	Yisha~Lu,
        Jiayu Liao,
        Xiang~Li,
        Hu~Huang*
        and~Bowen Zhang*
        
    \thanks{
This research is supported by National Nature science Foundation of china (No.62306184), Natural Science Foundation of Top Talent of SZTU (grant no. GDRC202320) and the Research Promotion Project of Key Construction Discipline in Guangdong Province (2022ZDJS112).}

\thanks{Fuqiang~Niu, Genan~Dai, Yisha~Lu, Xiang~Li, and Bowen Zhang are with the College of Big Data and Internet, Shenzhen Technology University, Shenzhen, 518000 China.}
\thanks{Jiayu Liao is affiliated with the University of Washington, Seattle, WA, USA.}
\thanks{Hu Huang is affiliated with the University of Science and Technology of China, Hefei, 518000 China.}

\thanks{Corresponding authors: Bowen Zhang and Hu Huang (e-mail: zhang\_bo\_wen@foxmail.com, huanghu@mail.ustc.edu.cn). }
\thanks{$^\dag$ These authors contributed equally to this work.}
}


\maketitle

\begin{abstract}

In the realm of contemporary social media, automatic stance detection is pivotal for opinion mining, as it synthesizes and examines user perspectives on contentious topics to uncover prevailing trends and sentiments. Traditional stance detection research often targets individual instances, thereby limiting its capacity to model multi-party discussions typical in real social media scenarios. This shortcoming largely stems from the scarcity of datasets that authentically capture the dynamics of social media interactions, hindering advancements in  conversational stance detection.
In this paper, we introduce MT$^2$-CSD, a comprehensive dataset for multi-target, multi-turn conversational stance detection. 
To the best of our knowledge, MT$^2$-CSD is the largest dataset available for this purpose, comprising 24,457 annotated instances and exhibiting the greatest conversational depth, thereby presenting new challenges for stance detection.
To address these challenges, we propose the Large Language model enhanced Conversational Relational Attention Network (LLM-CRAN), which exploits the reasoning capabilities of LLMs to improve conversational understanding. We conduct extensive experiments to evaluate the efficacy of LLM-CRAN on the MT$^2$-CSD dataset. The experimental results indicate that LLM-CRAN significantly outperforms strong baseline models in the task of conversational stance detection.

\end{abstract} 

\begin{IEEEkeywords}
Conversational Stance Detection, Knowledge Fusion, Attention Network.
\end{IEEEkeywords}

\section{Introduction}

Social media platforms have emerged as crucial venues for public discourse on contentious issues. These digital spaces offer valuable insights into public sentiment across a spectrum of topics, from healthcare policies to political debates~\cite{10537616}. This rich data landscape presents significant opportunities for computational analysis, particularly in web mining and content analysis. The extracted insights can inform decision-making processes across various domains, including marketing strategies and political campaigns~\cite{zhang2023twitter, 10.1145/3701716.3715307}. Consequently, automated stance detection in social media has become a critical component of opinion mining, facilitating a more nuanced understanding of user perspectives on diverse subjects.

\begin{figure*}
\centering
\includegraphics[width=0.9\linewidth]{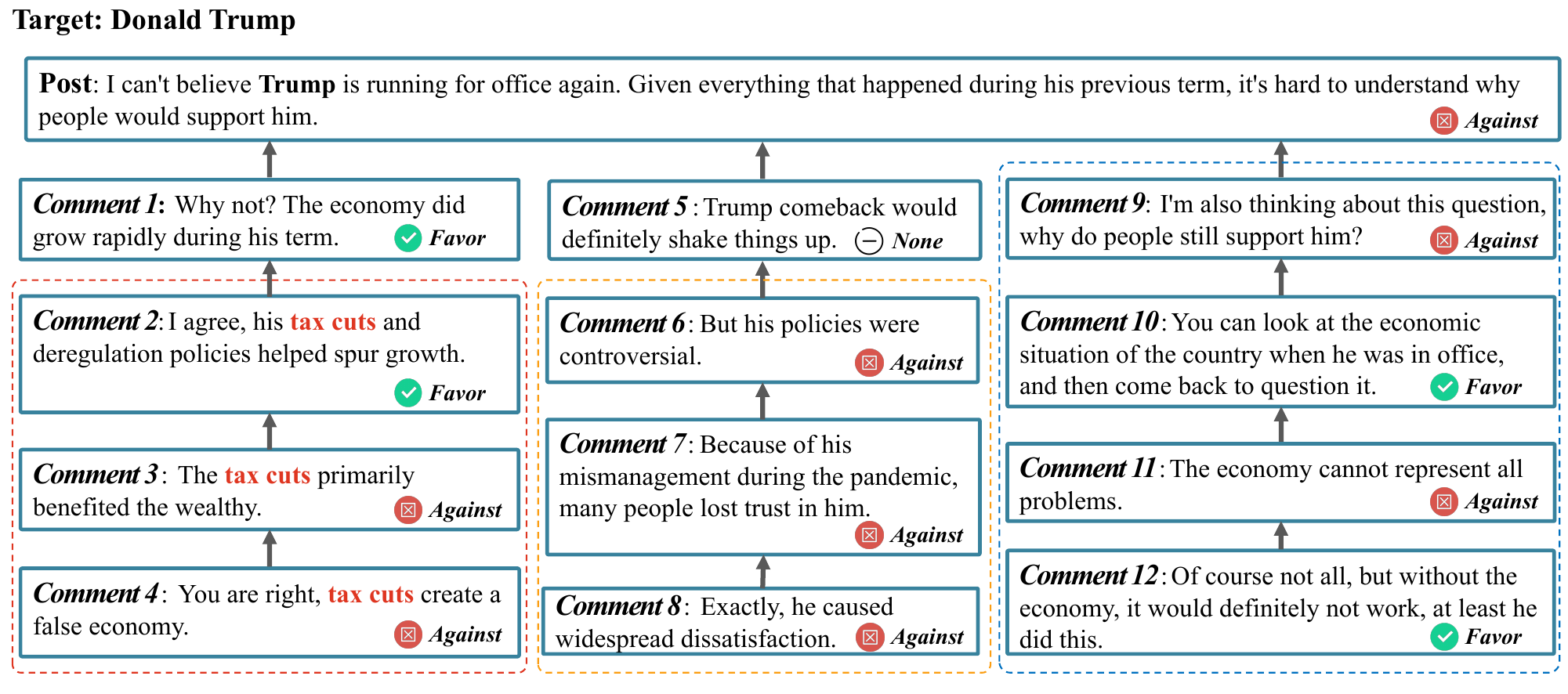}
\caption{An example of conversation on Reddit.} 
\label{fig:example}
\end{figure*}

The goal of stance detection is to identify the attitude or opinion expressed in text (e.g., statements, tweets, articles, or comments) towards specific targets~\cite{li2021p}.
Existing research typically categorizes this field into target-specific~\cite{li2023stance}, cross-target~\cite{ding2024cross}, and zero-shot~\cite{li2023tts} stance detection, primarily focusing on single-sentence analysis. 
However, in social media environment, users often express their views through conversational exchanges.
Methods that detect stances without considering context often struggle to make accurate predictions in conversation-based scenarios.
For instance, Fig.~\ref{fig:example} illustrates a social media discussion where comments center on the topic of remote work mentioned in the original post.
Accurate stance detection of individual comments necessitates a thorough analysis of the complete conversational context.
Consequently, Conversational Stance Detection (CSD) has gained increasing research attention, aiming to identify stances within conversation threads.

So far, three CSD datasets have been created to serve as benchmarks for the CSD task: 
SRQ~\cite{villa2020stance}, Cantonese-CSD (CANT-CSD)~\cite{li2022improved}, and CTSDT~\cite{li2023contextual}. 
However, these datasets have several drawbacks:
(1) Existing datasets primarily contain examples with few reply turns. For instance, the SRQ dataset includes only direct replies, representing single-turn comments. Similarly, in the CANT-CSD dataset, only 6.3\% of the data contains more than 3 reply rounds.
(2) The annotation quality of existing datasets is suboptimal. Specifically, the SRQ dataset annotates only the stance of reply texts, neglecting annotations for original comments. The CTSDT dataset relies mainly on automated annotations rather than manual labeling.
(3) The CANT-CSD dataset is restricted to Cantonese and has few annotated examples.
These problems limit the use of CSD models in real social media situations. Therefore, creating a high-quality CSD dataset is crucial for advancing research in this field.


To advance research in CSD, we introduce a new multi-target multi-turn conversational stance detection dataset, termed MT$^2$-CSD. This dataset comprises 24,457 meticulously annotated instances, representing a significant increase in scale compared to previous stance detection datasets.
In contrast to conventional datasets, MT$^2$-CSD boasts a high proportion of comments exceeding 4 turns, accounting for 67.47\% of the total. Notably, compared to CANT-CSD, the only other dataset containing multi-turn (beyond two turns) comments, MT$^2$-CSD offers 16 times more instances with a depth of 4 turns. 
This substantial growth offers a more thorough and varied set of conversational data for stance modeling.
The MT$^2$-CSD dataset introduces unique challenges for stance detection: 
(1) 
Implicit target mentions within local sub-discussions necessitate a detailed understanding of contextual information.
For example, in Fig.\ref{fig:example}, $Comment$ 2 to 4 discuss the impact of ``tax cut'' policy within a local discussion.
(2) While posts directly referencing the target provide explicit stance cues, determining the stance in comments necessitates a more intricate process involving the resolution of complex semantic relations within the text {and understanding the underlying context within the text. 
For example, in Fig.\ref{fig:example}, semantically, $Comment$ 6 contrasts with $Comment$ 5, $Comment$ 7 explains the reason behind $Comment$ 6's viewpoint, and $Comment$ 8 summarizes and expresses dissatisfaction with Trump. In terms of context, $Comment$ 9 aligns with the post's viewpoint and poses a question, $Comment$ 10 offers a suggestion to express its stance, and $Comment$ 11 and 12 express their viewpoints through rebuttals.}
In response, this paper aims to propose a stance detection method that integrates massive semantic knowledge to enhance conversational context learning. To achieve this, we focus on three key types of knowledge:
\textit{Local structure knowledge} representation facilitates the identification of discourse segments in conversational threads, enhancing coherence and reducing misinterpretation risks \cite{grosz1986attention}.
\textit{Logical relations} refer to the inferential connections between individual statements or events within a conversation \cite{fillmore2001frame}.
\textit{Conversation acts} denote the information and intentions conveyed or received by participants during the course of a conversation \cite{goodwin1990conversation}. 
The logical relations and conversation acts are illustrated in Table~\ref{tab:relation}.

\begin{table}
\centering
\caption{\label{tab:relation} Logical Relations and Conversation Acts classification.}
\fontsize{9pt}{12pt}\selectfont
  \begin{tabular}{>{\centering\arraybackslash}m{0.25\linewidth}>{\centering\arraybackslash}m{0.6\linewidth}}
    \hline
    \textbf{Relation} & \textbf{Classification} \\
    \hline
    \textbf{Logical Relations} & Contrastive, Succession, Causal, Summary\\
    \hline
 \textbf{Conversation Acts} & Summarize, Suggestion, Disagreement,Agreement, Refusal, Question, Clarification, Other\\
 \hline
  \end{tabular}
\end{table}

 \begin{table}
 \centering
 \caption{\label{tab:post-t} Differences between Specific Targets and Post-T}
\fontsize{9pt}{11pt}\selectfont
  \begin{tabular}{|>{\centering\arraybackslash}m{1cm}|>{\centering\arraybackslash}m{1cm}|>{\arraybackslash}m{5cm}|}
    \hline
    \textbf{Type}& \textbf{Target}& \textbf{Example}\\
    \hline
     \textbf{Specific-Target}& \textit{Donald Trump}& \textit{Post}: Trump's policies have made America great again. [stance: \textit{Favor}]  \newline
\textit{Comment 1}: Trump's policies have divided the nation, causing significant harm. [stance: \textit{Against}]\\
\hline
     \textbf{Post-T}& \textit{Post text} & \textit{Post}: I believe climate change is the biggest threat to humanity. \newline
\textit{Comment 1}: Climate change is exaggerated; real issue is economic instability. [stance: \textit{Against}]\\
     \hline
  \end{tabular}
\end{table}

\begin{table*}[htbp]
\centering
  \caption{\label{tab:datasets-details} Stance detection datasets}
\fontsize{9pt}{12pt}\selectfont
    \begin{tabular}{>{\raggedright\arraybackslash}m{0.1\linewidth}
    >{\raggedright\arraybackslash}m{0.15\linewidth} 
    >{\raggedright\arraybackslash}m{0.55\linewidth} 
    >{\centering\arraybackslash}m{0.05\linewidth} 
    >{\centering\arraybackslash}m{0.05\linewidth}}
    \hline
     \textbf{Type}&\textbf{Dataset-name} & \textbf{Target(s)} & \textbf{Size} & \textbf{Time} \\
    \hline
     \multirow{5}{0.08\linewidth}{\textbf{Sentence level}}&
     Sem2016 & Atheism, Climate Change is Concern, Feminist Movement, Hillary Clinton, Legalization of Abortion, Donald Trump& 4,870& 2016 \\
     &P-stance & Donald Trump, Joe Biden, Bernie Sanders & 21,574& 2020 \\
     &WT-WT & Aetna, Express Scripts, Cigna, Humana, 21st Century Fox & 51,284& 2020 \\
     &VAST & Various topics & 23,525& 2020 \\
     &COVID-19-Stance & Anthony S. Fauci, M.D., Keeping Schools Closed, Stay At Home Orders, Wearing a Face Mask & 7,122& 2020 \\
     \hdashline
     \multirow{4}{0.08\linewidth}{\textbf{Conversation  level}}&
    CSD& Covid-19 & 5,876& 2022 \\
  &SRQ & Student Marches, Iran Deal, Santa Fe Shooting, General Terms & 5,220&2020 \\
     &CTSDT& Covid-19 & 50,467 & 2023 \\
     &
 \textbf{MT$^2$-CSD}& Bitcoin, Tesla, SpaceX, Donald Trump, Joe Biden, Post-T&24,457 &2024\\
    \hline
  \end{tabular}
\end{table*}

Traditionally, acquiring such knowledge requires extensive manual annotation. 
In contrast, Large Language Models (LLMs), such as the GPT series, excel in text comprehension and generation due to their pre-training on massive, knowledge-rich corpora. These capabilities make LLMs a promising alternative to manual annotation, enabling automatic extraction of complex relational knowledge from text.
To leverage these capabilities, in this paper, we propose the LLM-based Conversational Relational Attention Network ({LLM-CRAN}), utilizing LLM to extract logical relations and conversation acts within conversations.
The LLM-CRAN framework comprises two primary components: an LLM-driven Knowledge Acquisition Module (KAM) and a Multi-Knowledge Integrated Attention Network (MKIAN). KAM employs a zero-shot prompting approach, constructing templates based on conversation content to elicit the relational knowledge mentioned above from conversations.
MKIAN, designed to integrate multiple knowledge types, consists of four main layers.
Local Knowledge Layer: Implements Convolutional Neural Networks (CNN) with a masking mechanism to identify local paragraphs within the complete conversation.
Contextual Representation Layer: Utilizes Graph Convolutional Networks (GCN) to model context information based on comment dependency relations.
Logical Relation Layer and Conversation Act Layer: encode Logical and conversation act knowledge using Relational Graph Convolutional Networks (RGCN).
Finally, we employ a multi-hop attention network to effectively fuse various types of knowledge.

This work substantially extends our previous research~\cite{niu2024challenge}. We present two key extensions. First, we expanded the dataset from 15,876 to 24,457 instances and introduced a new stance detection target based on posts (Post-T). Table~\ref{tab:post-t} illustrates the distinctions between Post-T and the original target-based approach. Post-T analyzes comment stances towards opinions expressed in posts, while the target-based method examines post and comment stances on specific targets. 
Second, we introduce LLM-CRAN, which effectively expands upon the original version's knowledge integration capacity. Experimental comparisons with several robust baseline methods demonstrate that LLM-CRAN significantly outperforms current state-of-the-art approaches.

In this article, the main contributions of our work can be summarized as follows:
\begin{enumerate}{}{}
\item  We present a challenging multi-target multi-turn conversation stance detection (MT$^2$-CSD) dataset designed for conversational stance detection. 
This is the largest English conversational stance dataset labeled by humans so far. 
Releasing MT$^2$-CSD will advance research in stance detection.

\item We propose LLM-CRAN, a novel model that effectively integrates multiple semantic knowledge types, enhancing the model's ability to model complex semantic relations.
\item 
We thoroughly evaluate the state-of-the-art stance detection methods using widely adopted approaches.
Our experimental findings illuminate the challenges faced by current models in CSD. Notably, extensive experiments demonstrate that our knowledge-augmentation framework significantly enhances both accuracy and generalization in stance detection tasks.
\end{enumerate}

The paper is structured as follows:
Section 2 presents a comprehensive review of traditional and recent stance detection approaches.
Section 3 details our proposed dataset MT$^2$-CSD.
Section 4 describes our proposed framework LLM-CRAN.
Section 5 reports and analyzes the experimental results.
Section 6 concludes with key findings and directions for future research.

\section{Related Work}

\subsection{Stance Detection Datasets}
\paragraph{Sentence-level stance detection datasets}
To date, several benchmark datasets for stance detection in social media have been developed and widely adopted by the research community. 
The features of these datasets are shown in Table ~\ref{tab:datasets-details}. 
SemEval-2016 Task 6 (SEM16) is the first stance detection dataset from Twitter and is a well-known benchmark, containing 4,870 tweets that express stances on different targets \cite{MohammadKSZC16}.
Later, to take advantage of large annotated datasets, Zhang et al.\cite{zhang2020enhancing} expanded SEM16 by adding the \textit{Trade Policy} target. Conforti et al.\cite{conforti2020will} also added to the WT-WT dataset by providing a larger labeled collection.
Furthermore, Li et al.\cite{li2021p} introduced the P-Stance dataset, aimed at the political domain, featuring tweets with longer average lengths. Glandt et al.\cite{glandt2021stance} created a dataset for COVID-19 stance detection. In addition to these stance detection datasets focused on specific targets, Allaway et al.~\cite{allaway2020zero} proposed the VAST dataset, which focuses on zero-shot stance detection with over a thousand different targets. These efforts mainly concentrate on sentence-level (individual post-level) stance detection tasks.

\paragraph{CSD datasets}
At present, there are only three CSD datasets designed specifically for comments in conversation threads.
The SRQ dataset~\cite{villa2020stance} was created to handle stance detection in tweet replies and quotes.
However, it focuses only on single-turn replies and quotes. 
The CANT-CSD dataset~\cite{li2022improved} is created for stance detection in multi-turn conversations. Even though it covers a lot, most of the data in CANT-CSD is limited to short reply turns. Specifically, 80.1\% of the data has only two turns of replies, and just 6.3\% has more than three turns. We've noticed that, especially with trending topics, the depth of comment replies often goes beyond five turns, which limits CANT-CSD's use in real-world cases. Additionally, the CANT-CSD dataset is annotated in Cantonese, reducing its broader impact in the stance detection field.
The CTSDT dataset~\cite{li2023contextual} includes deeper reply rounds; however, it relies on automated annotation, which raises concerns about the quality of the annotations.

Given the limitations of existing CSD datasets, the MT$^2$-CSD addresses the limitations of existing CSD datasets by providing a larger proportion of multi-turn conversations and detailed annotations for complex conversational dynamics.

\begin{table*}
\normalsize
  \caption{ \label{tab:Sources} Data sources for MT$^2$-CSD}
\centering
\fontsize{9pt}{12pt}\selectfont
  \begin{tabular}{>{\centering\arraybackslash}m{0.1\linewidth}>{\raggedright\arraybackslash}m{0.8\linewidth}}
    \hline
    \textbf{Target} & \multicolumn{1}{c}{\textbf{Subreddits/Topics}} \\
    \hline
    \textbf{Bitcoin} & {r/Bitcoin, r/BitcoinBeginners, r/BitcoinMarkets, r/CryptoCurrency, r/btc, r/BitcoinMining, r/BitcoinCA, r/bitcoin\_uncensored, r/Bitcoincash, r/therewasanattempt}\\
    \hline
    \textbf{Tesla} & {r/Tesla, r/TeslaModel3, r/TeslaLounge, r/teslamotors, r/RealTesla, r/TeslaModelY, r/teslainvestorsclub, r/TeslaPorn, r/teslainvestorsclub, r/electricvehicles, r/TeslaCam, r/therewasanattempt} \\
    \hline
    \textbf{SapceX} & {r/spacex, r/elonmusk,  r/Spaceexploration, r/space, r/Starlink, r/technology, r/SpaceXLounge, r/Space\_\_X, r/SpaceXStarship}\\
    \hline
    \textbf{Biden} & {r/JoeBiden, r/BidenWatch, r/Conservative, r/politics, r/conspiracy, r/news, r/WhatBidenHasDone, r/Republican, r/democrats} \\
    \hline
    \textbf{Trump} & {r/trump, r/TrumpCriticizesTrump, r/politics, r/AskTrumpSupporters, r/news, r/Conservative, r/Republican, r/AskThe\_Donald, r/democrats}\\
    \hline
    \textbf{Post-T} & t/Animals and Pets, t/Anime, t/Art, t/Cars and Motor Vehicles, t/Crafts and DIY, t/Culture, t/Race and Ethnicity, t/gaming,t/sport, t/business,t/relevision\\
    \hline
  \end{tabular}
\end{table*}
\subsection{Stance Detection Approaches}

\paragraph{Approaches Based on Deep Neural Networks}
The evolution of deep learning techniques in the domain of stance detection has seen significant developments, each uniquely enhancing the field. The LSTM~\cite{hochreiter1997long} have demonstrated exceptional results in stance detection by effectively managing long-term dependencies in textual data~\cite{zarrella-marsh-2016-mitre}. Alongside LSTM, recurrent neural networks (RNN) and CNN have also been extensively utilized. 
Numerous studies~\cite{igarashi-etal-2016-tohoku, zhou2017connecting, wei2019topic} demonstrate the effectiveness of RNN and CNN in extracting spatial hierarchies of features, proving highly effective for processing structured text data.

Building on these foundational techniques, attention-based methods have been developed. These methods utilize target-specific information as the attention query, deploying an attention mechanism to infer stance polarity~\cite{dey2018topical, du2017stance, sun2018stance}. Furthermore, GCN have been widely applied to model the relationships between targets and output texts, extending the capabilities of convolutional networks~\cite{LiPLSLWYH22, ConfortiBPGTC21,zhang2025logic}.

\paragraph{Approaches Based on Pre-trained Models}
With the advent of pre-trained language models (PLMs), the application of BERT~\cite{devlin-etal-2019-bert} (Bidirectional Encoder Representations from Transformers) through fine-tuning has demonstrated remarkable success in capturing nuanced semantic relations, which are crucial for accurately discerning subtle differences in stance. Several representative methods perform fine-tuning with PLMs as baselines. The pre-trained BERT is fine-tuned on the training data, leveraging its robust language understanding capabilities. JoinCL~\cite{liang2022jointcl} uses stance contrastive learning and target-aware prototypical graph contrastive learning for stance detection. It is designed to apply stance features to new, unseen targets, improving the model's flexibility. TTS~\cite{li2023tts} uses target-based data augmentation to find key targets in each training sample and then uses these augmented targets for zero-shot stance detection.

In addition to fine-tuning approaches, prompt-tuning methods with PLMs have also been explored. KPT~\cite{shin2020autoprompt} introduces external lexicons to define the verbalizer, utilizing SenticNet instead of sentiment lexicons for more contextually appropriate prompts. KEPrompt~\cite{KEPrompt} employs an automatic verbalizer to define the label words automatically, reducing the need for manual intervention and improving scalability. These diverse methodologies highlight the progressive integration of PLMs through both fine-tuning and prompt-tuning techniques, collectively enhancing the accuracy and robustness of stance detection models.

\paragraph{Approaches Based on Large Language Models}

Advancements in LLM technologies have considerably enhanced methods of background knowledge augmentation due to the rich contextual understanding they possess. One prominent approach treats LLMs as efficient knowledge repositories, harnessing their stored information through targeted prompting techniques~\cite{zhang2022would, zhang2023investigating}. For example, Cai et al.~\cite{cai2023human} proposed a human-in-the-loop system augmented with CoT prompting, investigating how manual correction of sub-logic in rationales can refine LLMs reasoning. 
Another approach combines LLMs with traditional methods~\cite{zhang2024surveystancedetectionsocial}. The first phase involves decomposing extensive text corpora into segments within a dedicated information retrieval system. The second stage integrates the retrieved knowledge with input prompts before feeding them into trainable models~\cite{lan2023stance, li2023stance,ding2024cross,upadhyaya2025interpretable}. 
In addition, recent work has explored solving stance detection tasks by fine-tuning large-scale multimodal models~\cite{niu2024multimodal}.

\section{MT$^2$-CSD Dataset}
This section details the development process of our MT$^2$-CSD\footnote{https://github.com/nfq729/MT-CSD} dataset, which consists of 24,457 texts extracted from Reddit.




\begin{table}[hbtp]
\centering
\caption{\label{tab:post} The number of data items for each target.}
\fontsize{9pt}{12pt}\selectfont
\resizebox{\linewidth}{!}{
\begin{tabular}{lcccccc}
\hline
\textbf{Type} & \textbf{Bitcoin} & \textbf{Tesla} & \textbf{SpaceX} & \textbf{Biden} & \textbf{Trump} & \textbf{Post-T} \\
\hline
\textbf{Post} & 93 & 52 & 32 & 72 & 81 & 374 \\
\textbf{Comment} & 9,716 & 8,989 & 4,911 & 10,593 & 10,203 & 42,536 \\
\hline
\end{tabular}
}
\end{table}

\subsection{Data Collection}

To obtain authentic social media conversational data, we leveraged Reddit\footnote{https://www.reddit.com}, one of the largest online forums, ensuring the richness and authenticity of the collected MT$^2$-CSD dataset.
We accessed the data through Reddit's official API\footnote{https://www.reddit.com/dev/api}. 
To obtain topics with sufficient discussion and high relevance, during the data collection process, we gathered Reddit posts and associated popularity metrics, such as upvotes and comment counts. 
A manual review was conducted to assess the relevance of the posts to the given targets, ensuring that the collected posts were highly pertinent and featured sufficiently in-depth comments to facilitate dataset annotation. 
Subsequently, we collected comments for each selected post. The resulting dataset encompassed relevant posts, associated discussions, and comments, providing a comprehensive overview of conversations centered around the specified targets. The selected targets for this dataset included ``\textit{Tesla}'', ``\textit{SpaceX}'', ``\textit{Donald Trump}'', ``\textit{Joe Biden}'', and ``\textit{Bitcoin}'', and we additionally constructed posts as targets (Post-T). The sources of the data are reported in Table~\ref{tab:Sources}.


\begin{table}[htbp]
\centering
\caption{\label{tab:consistency} Annotation consistency and agreement.}
 \fontsize{9pt}{12pt}\selectfont
  \begin{tabular}{ccc}
    \hline
    \textbf{Target} & \textbf{kappa} & \textbf{consistency} \\
    \hline  
    \textbf{Bitcoin} & 0.79& 0.93\\
 \textbf{Tesla} & 0.75 & 0.74 \\
 \textbf{SpaceX} & 0.79 & 0.83 \\
 \textbf{Biden} & 0.71& 0.96 \\
 \textbf{Trump} & 0.74 & 0.71 \\
 \textbf{Post-T}& 0.65& 0.76\\
 \hline
    \textbf{Avg.} & 0.74& 0.82\\
    \hline
  \end{tabular}
  
\end{table}

\subsection{Data Preprocessing}
To maintain the high quality of the MT$^2$-CSD dataset, we applied several strict preprocessing steps:
\begin{enumerate}{}{}
    \item High Relevance to Target: Each post's content must be highly relevant to the specified target. A two-reviewer process was used to evaluate this relevance, retaining only posts considered highly relevant by both reviewers.
    \item Minimum 200 Comments per Post: To ensure substantial attention and discussion for each post, we required at least 200 comments per post. Posts with fewer comments would lack sufficient conversation depth and complexity.
    
    \item Appropriate Text Length: We set limits on the length of posts. To ensure quality, posts had to be at least 15 words but no more than 150 words. Texts with fewer than 15 words are either too simple for stance detection or too noisy, while posts longer than 150 words often have repeated expressions.
    
    \item Excluding Non-English Posts: To build an all-English dataset, non-English posts were systematically excluded to maintain language consistency. Multilingual stance detection is considered a potential area for future research.
    
\end{enumerate}
After this rigorous data filtering process, the resulting data distribution is summarized in Table~\ref{tab:post}.

\subsection{Data Annotation and Quality Assurance}
We developed an annotation system to ensure annotators thoroughly reviewed the preceding context and provided accurate attitude labels. This system is designed for conversational data and aims to streamline and improve the data annotation process. During annotation, clear guidelines were given to annotators, instructing them to label each comment as ``\textit{against}'', ``\textit{favor}'', or ``\textit{none}'' to indicate the attitude. Additionally, annotators were asked to specify whether newly added comments were related to the specified target.

We invited eleven researchers with expertise in natural language processing to annotate the data. Before the formal annotation process, we conducted two pilot annotation rounds to ensure the reliability of the annotated data. Three additional expert annotators reviewed the pilot annotations to confirm that each annotator could effectively perform the task. During the formal annotation stage, each data instance was labeled by at least two annotators. In cases of disagreement between the initial two annotators, an additional annotator was involved in labeling the contentious statements, and a final consensus was reached through voting. This approach not only ensured the reliability of the data but also integrated inputs and consensus from multiple annotators, enhancing the overall quality of the stance labels assigned to each instance.

After obtaining the annotation results, we calculated the kappa statistic~\cite{kappa} and inter-annotator agreement to measure the consistency between annotators. Following \cite{li2021p}, we used the ``\textit{Favor}'' and ``\textit{Against}'' classes to compute the kappa values. The results are shown in Table~\ref{tab:consistency}. The kappa statistic achieved high scores, with an average of 74\%. The average inter-rater consistency, where agreement was rated as 1 and disagreement as 0, was 82\%. These results confirm that our dataset is well-annotated and of high quality.

\begin{table}
\fontsize{9pt}{12pt}\selectfont
\begin{center}
\caption{Label distribution of the MT$^2$-CSD dataset.}
\label{tab:Data distribution}
\begin{tabular}{cccccccc}
\hline
 \multirow{2}{*}{\textbf{Target}}& \multicolumn{7}{c}{\textbf{Samples and Proportion of Labels}} \\
 \cline{2-8} 
 & Against& \%& Favor& \%& None&\%  &Total
 \\
  \hline
 \textbf{Bitcoin}& 1,324& 39& 869& 26& 1,192& 35&3,385\\
 \textbf{Tesla}& 1,146& 31& 477& 13& 2,068& 56& 3,691\\
  \textbf{SpaceX}& 298& 15& 595& 29& 1,162& 57& 2,055\\
 \textbf{Biden}& 352& 12& 1,186& 40& 1,409& 48& 2,947\\
 \textbf{Trump}& 1,667& 44& 207& 5& 1,924& 51&3,798\\
 \textbf{Post-T}& 2,619& 30& 3,505& 41& 2,457& 29&8,581\\
 \hline
 \textbf{Total}& 7,406& 30& 6,839& 28& 10,212& 42&24,457
\\
\hline
 \end{tabular}
 \end{center}
\end{table}

\subsection{Data Analysis}
Table \ref{tab:Data distribution} presents the statistics of our MT$^2$-CSD dataset. The final annotated dataset contains 24,457 instances, which is 4.2 times larger than the CANT-CSD and 5.2 times larger than the SRQ datasets. Table \ref{tab:statistics} shows the distribution of instances across different depths. A significant portion, 67.47\%, of the data in our MT$^2$-CSD dataset has a depth greater than 3. In contrast, only 6.3\% of the CANT-CSD dataset exceeds depth 3. We then divided the dataset into training, validation, and test sets for all targets in a 65/15/20 ratio, ensuring balanced representation for comprehensive evaluation and analysis.



\section{Methods}
In this section, we present a detailed description of our proposed LLM-based Conversational Relational Attention Neural Network (LLM-CRAN) model for conversational stance detection.

\subsection{Task Definition}

We represent the concatenation of full conversations as $C=[x_1, \ldots, x_n]$, where $[x_1, \ldots, x_{n-1}]$ denotes the conversation history and $x_n$ represents the current text whose stance is to be identified.
Let $X = \{C_i, t_i\}_{i=1}^{N}$ be the labeled dataset, where $t$ denotes the corresponding target. The goal of CSD is to determine the stance label $y$ of $x_n$ from $C_i$ with respect to the corresponding target $t$.

\subsection{Framework Overview}
\begin{figure*}[htbp]
\centering
\includegraphics[width=0.9\linewidth]{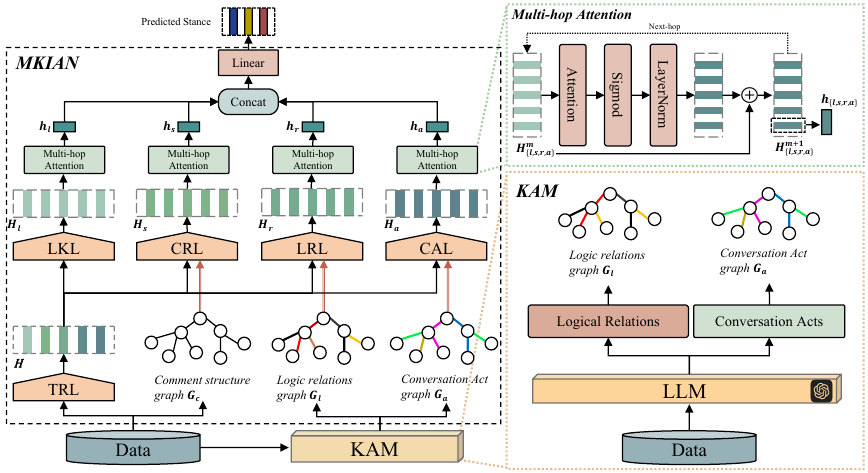}
\caption{The architecture of our LLM-CRAN framework.} 
\label{fig:model}
\end{figure*}

As illustrated in Fig.~\ref{fig:model}, the LLM-CRAN framework comprises two primary components: a KAM and MKIAN. 
KAM employs a zero-shot prompting approach, initially constructing natural language-based prompt templates. These templates are then input to the LLM as instructions, eliciting knowledge about logical relations and conversational acts.
MKIAN is designed to integrate multiple knowledge types and consists of four main layers.
Text Representation Layer: Utilizes BERT to generates a contextualized representation for each token in the input conversation text.
Local Knowledge Layer: Implements a CNN with a masking mechanism to identify local segments within the complete conversation.
Contextual Representation Layer: Utilizes GCN to model contextual information based on comment dependencies.
Logical Relations and Conversational Acts Layers: Employ RGCN to encode logical and conversational act knowledge.
Finally, a multi-hop attention network is utilized to effectively fuse various types of knowledge.

\begin{table}
\begin{center}
  \caption{\label{tab:statistics} Statistics of the MT$^2$-CSD dataset. Here, WC is short for word count.}
   \fontsize{9pt}{12pt}\selectfont
  \begin{tabular}{cccc}
    \hline
    \textbf{Instance} & \textbf{Avg. WC} & \textbf{Depth} & \textbf{Number}\\
    \hline
    \textbf{Post} & 36.63& 1 & 440 (1.80\%) \\
    \hline
    \multirow{7}{*}{\textbf{Comment}}
     & 32.09& 2 & 2,553 (10.44\%)\\
     & 29.45& 3 & 4,962 (20.29\%)\\
     & 31.83& 4 & 5,291 (21.63\%)\\
     & 32.79& 5 & 4,667 (19.08\%)\\
     & 34.27& 6 & 3,671 (15.01\%)\\
     & 35.44& 7 & 1,900 (7.77\%)\\
     & 38.33& 8 & 973 (3.98\%)\\
    \hline 
  \end{tabular}
\end{center}
\end{table}

\subsection{KAM}

As mentioned above, we extract logical relations and conversation acts from LLM by constructing prompt templates. Specifically, for the logical relation, following  \cite{li2022survey}, we define the relations between sentences in a conversation into four categories based on logical relations: ``contrastive'', ``succession'', ``causal'', and ``summary''. 
For conversation acts, we select seven high-frequency conversation acts from the 42 types listed in the literature~\cite{stolcke2000dialogue}, and include an additional ``other'' category, resulting in a total of eight categories. 
The complete set of categories is presented in Table~\ref{tab:relation}.
Formally, we define the prompt templates for logical relations and conversational acts as follows:

\begin{center}
    \fcolorbox{black}{darkgreen!40}{\parbox{0.95\linewidth}{\textsc{User:} 
The following are conversation threads, where ``Post'' content is considered a post on social media. Each comment is a reply to the preceding comment, and all comments are responses to the original post. [\textbf{Conversation Thread}]. Please analyze the relations between each post and comment, as well as between each comment. Determine the logical relation between \textbf{{$Comment \ x_i$}} and \textbf{{$Comment \ x_{i-1}$}}, selecting from [\textbf{Defined Logical Relations (or Conversation Acts)}].}}
\end{center}


In these templates, ``Conversation Thread'' refers to the entire conversation process above the $Comment \ x_i$. By using these prompt templates, we obtain the logical relation and conversation act of $Comment \ x_i$ with respect to $Comment \ x_{i-1}$ or the post. Through these steps, we obtain the logical relations $\mathcal{R}_l$ and conversation acts $\mathcal{R}_a$ of the full conversation.

\subsection{MKIAN}

\paragraph{Text Representation Layer (TRL)}
We utilize BERT to generate deep contextualized representations for input conversation. Formally, we first concatenate all sentences in $C$ into a token sequence $S_x$ in which every two consecutive sentences are separated with a special token $[SEP]$:
\begin{equation}
S_x  = [CLS] \ x_1 \  [SEP] \  x_2 \ [SEP]\  \ldots \ x_n \ [SEP]
\end{equation}

Subsequently, we append the stance detection target $w_t$ to the end of each sentence $x_i$, forming $S$:
\begin{equation}
S = [CLS] \ x_1 \ w_t \  [SEP] \  x_2 \ w_t \ [SEP]\  \ldots \ x_n \ w_t  \ [SEP]
\end{equation}

We then employ a BERT model to transform $S$ into an embedding representation, denoted as $E$. Next, we derive a vector representation for each sentence, $h_{x_i}$, by computing the mean of its constituent word vectors. 
The sentence vectors for the entire conversation, $H \in \mathbb{R}^{N \times D}$, are then obtained by concatenating and aggregating the individual sentence vectors $h_{x_i}$.


After obtaining the sentence embeddings $H$, the MKIAN performs computations across four layers. Each layer generates a sentence embedding vector, and a common multi-hop attention operation is implemented to integrate the representations from all four layers. We first provide an overview of the process for acquiring various sentence embedding vectors from each module. Subsequently, we offer a detailed description of the shared multi-hop attention operation.

\paragraph{Local Knowledge Layer (LKL)}
To effectively identify local information for the target comment $x_n$, we first pass $H$ through two one-dimensional convolutional layers with a kernel size of $\gamma$, operating from $x_1$ to $x_n$. Subsequently, we construct a binary mask vector, setting the $n$-th dimension to 1 and all other positions to 0. 
By applying this mask, we obtain a representation of the local segment corresponding to the kernel size, denoted as $H_l \in \mathbb{R}^{N \times D}$. These vectors maintain the same dimensionality as the original sentence vectors but are enriched with local information.


\paragraph{Contextual Representation Layer (CRL)}
The contextual representation layer models the complete conversation content based on the reply relations between comments. First, we construct a comment graph $G_c = (V, E)$ from the conversation history, where nodes $V$ represent sentence vectors $h_{x_i}$, and edges $E$ denote comment relations. After obtaining the sentence embedding vectors $H$, we input them into a two-layer GCN. The graph representation $H_s \in \mathbb{R}^{N \times D}$ is computed as follows:


\begin{equation}
\label{1}
H_s =\sigma(A \sigma ({A}HW_0)W_1)
\end{equation}
where $A = A + I_N$ is the adjacency matrix of the undirected graph $G_c$ with added self-connections. $I_N$ is the identity matrix, $W_0$ and $W_1$ are trainable parameters. $\sigma$ denotes an activation function.

\paragraph{Logical Relations Layer (LRL)}
In the logical relations layer, we first formalize the logical relations obtained from the KAM as a subgraph. 
This is represented as $G_l = (V, E, \mathcal{R}_l)$, where $\mathcal{R}_l$ denotes the types of logical relations within the conversation thread. 
Subsequently, we construct an RGCN to model and represent $G_l$. 
Formally, after obtaining the sentence embedding vectors $H$ and relations $\mathcal{R}_l$, we input them into the RGCN. 
The output of $l+1$ layer is calculated as follows:


\begin{equation}
h_{x_i}^{r,(l+1)}=\sigma\left(\sum_{\zeta \in \mathcal{R}_l} \sum_{j \in \mathcal{N}_{i}^{\zeta}} \frac{1}{c_{i, \zeta}} W_{\zeta}^{r,{(l)}} h_{x_j}^{r,(l)}+W_{0}^{r,{(l)}} h_{x_i}^{r,(l)}\right)
\end{equation}

where $\mathcal{N}_{i}^{\zeta}$ denotes the set of neighbors of node $i$ under relation $\zeta \in \mathcal{R}_l$. The term $c_{i, \zeta}$ is a normalization constant specific to the problem, defined as $|\mathcal{N}_{i}^{\zeta}|$. Here, $h_{x_j}^{r,(l)}$ and $h_{x_i}^{r,(l)}$ are the sentence embedding vectors, and $W_{\zeta}^{r,{(l)}}$ and $W_{0}^{r,{(l)}}$ are trainable parameters. The function $\sigma$ denotes an activation function.
After two layers of the RGCN, we obtain the final sentence embeddings $H_r = \{h_{x_i}^{r,(3)}|i\in (1,\ldots,n)\}$, where $H_r \in \mathbb{R}^{N \times D}$.

\paragraph{Conversation Act Layer (CAL)}
The construction of the conversational act layer is analogous to that of the logical relations layer. We begin by formulating the conversational act relations obtained from the KAM into a graph $G_a = (V, E, \mathcal{R}_a)$, where $\mathcal{R}_a$ represents the types of conversational acts within the conversation thread. 
Subsequently, after obtaining the sentence embedding vectors $H$ and relations $\mathcal{R}_a$, we input them into an RGCN. 
The output of $l+1$ layer is calculated as follows:
\begin{equation}
h_{x_i}^{a,(l+1)}=\sigma\left(\sum_{\zeta \in \mathcal{R}_a} \sum_{j \in \mathcal{N}_{i}^{\zeta}} \frac{1}{c_{i, \zeta}} W_{\zeta}^{a,{(l)}} h_{x_j}^{a,(l)}+W_{0}^{a,{(l)}} h_{x_i}^{a,(l)}\right)
\end{equation}
where $\mathcal{N}_{i}^{\zeta}$ denotes the set of neighbors of node $i$ under relation $\zeta \in \mathcal{R}_a$. The term $c_{i, \zeta}$ is a normalization constant specific to the problem, defined as $|\mathcal{N}_{i}^{\zeta}|$. Here, $h_{x_j}^{a,(l)}$ and $h_{x_i}^{a,(l)}$ are the sentence embedding vectors, and $W_{\zeta}^{a,{(l)}}$ and $W_{0}^{a,{(l)}}$ are trainable parameters. The function $\sigma$ denotes an activation function.\
After two layers of the RGCN, we obtain the final sentence embeddings $H_a = \{h_{x_i}^{a,(3)}|i \in (1, \ldots , n)\}$, where \(H_a \in \mathbb{R}^{N \times D}\).

\paragraph{Multi-hop Attention}


Following the acquisition of sentence vectors ($H_l$, $H_s$, $H_r$, $H_a$), let $H_{\{l,s,r,a\}}^0 = H_{\{l,s,r,a\}}$. These vectors serve as inputs for a multi-hop attention mechanism. For each iteration $m$, the sentence vectors are utilized in the following computations:
\begin{equation}
R_{\{l,s,r,a\}}^m = c_{\{l,s,r,a\}}^m H_{\{l,s,r,a\}}^m \ \
\end{equation}
where $c_l^m, c_s^m, c_r^m, c_a^m$ (dimensions $\mathbb{R}^{N \times 1}$) are obtained by multiplying the last sentence vector $h_{x_n} \in \mathbb{R}^{1 \times D}$ and the sentence embedding vectors ($H_l^m$, $H_s^m$, $H_r^m$, $H_a^m$).

Subsequently, these vectors undergo transformation through an activation function and layer normalization. The resulting outputs are then scaled by a factor $\lambda$ and added to the previous sentence embedding vectors.
\begin{gather}
    H^{m+1}_{\{l,s,r,a\}} = \lambda LN(\sigma(R_{{\{l,s,r,a\}}}^{m})) + H_{{\{l,s,r,a\}}}^{m}
\end{gather}
where $LN$ represents layer normalization, and $\sigma$ denotes the sigmoid activation function. We repeat the Multi-hop Attention module $p$ times to obtain the final sentence vectors.

After $p$ iterations of the aforementioned process, we extract the final vector from each set ($H_l^p$, $H_s^p$, $H_r^p$, $H_a^p$), each with dimensions $\mathbb{R}^{1 \times D}$. These resultant sentence vectors are denoted as $h_l$, $h_s$, $h_r$, and $h_a$, respectively.

The predicted stance $\hat{y}$ is then computed using a fully connected layer.
For stance detection, we employ the cross-entropy between the predicted stance $\hat{y}$ and the ground-truth stance $y$ as our loss function:

\begin{gather}
\mathcal{L} = - \sum_{i=1}^{N} \sum_{j=1}^{C} y_{ij} \log \hat{y}_{ij},
\end{gather}
where $N$ represents the number of instances in the training set. $C$ denotes the number of possible stance categories.

\begin{table*}
\begin{center}	
\normalsize
\caption{\label{tab:result} The performance of baseline models for in-target stance detection on the six targets in the MT$^2$-CSD dataset. In the experiments, ``Avg.'' represents the average of all targets' $F_{avg}$ scores. Bold indicates the highest score, and underline indicates the second-highest score.} 
\fontsize{9pt}{12pt}\selectfont
    \begin{tabular}{l|ccccccc}
    \hline
    \textbf{Methods}& Bitcoin& Tesla& SpaceX& Biden& Trump  &Post-T& Avg.\\  
    \hline
    \multicolumn{8}{c}{Only considering individual posts/comments}\\ 
    \hline
BiLSTM        & 32.99   & 31.40 & 22.79& 25.54 & 24.47  &30.94&28.02
\\
TAN            & 33.68 &  33.19   & 25.86 & 26.43 & 25.84 &27.59& 28.77
\\
GCAE       & 46.25  & 36.70   & 38.37 & 25.42 & 35.34  &35.63&36.29
\\
CrossNet       & 32.73  & 31.76   & 30.12 & 20.28 & 30.27 &35.72&30.15
\\
\hdashline
MPT       & 49.45  & 41.80  & 46.38 & 27.98 & 36.50  &42.27& 40.73
\\
KPT            & 50.34 &  43.11   & 47.47 & 28.90 & 41.87  &40.13&41.97
\\
KEPrompt            & 50.34  & 41.23 & 47.11 & 30.31 & 40.87 &39.68&41.59
\\ 
\hdashline
BERT       & 50.99   & 43.72 & 45.88 & 26.65 & 42.45 &41.85&41.92
\\
TTS            & 50.88  & 43.85 & 47.50 & 29.00 & 42.10  &34.96&41.38
\\ 
JoinCL            & 50.21  & 31.06 & 51.47 & 26.32 & 34.54  &37.25&38.48
\\
\hline
\multicolumn{8}{c}{Considering conversation history}\\ 
\hline
BiLSTM         & 44.27  & 35.55  & 28.15    & 27.36  & 26.47   &34.25& 32.68
\\
TAN             & 40.78  & 39.31  & 28.15    & 28.35  & 29.31  &30.53& 32.74
\\
GCAE       & 48.75 & 42.75 & 42.07  & 30.10  & 39.43 &40.32& 40.57
\\
CrossNet       & 37.73 & 31.76 & 33.63  & 25.49 & 37.94 &38.23& 34.13
\\
\hdashline
MPT         & 51.42   & 44.53   & 51.30     & 31.08   & 38.84   &44.51& 43.61
\\  
KPT            & 53.22 & 46.67& 52.65   & 32.22& 43.97 &42.74& 45.25
\\
KEPrompt                & 53.22     & 45.64      & 50.91       & 31.08    & 43.64     &45.92& 45.07
\\  
\hdashline
BERT       & 53.60 & 47.39 & 49.31   & 29.13 & 45.11&46.53& 45.18
\\
TTS               & 53.60& 46.08      & 52.41       & 31.23     & 44.41 &38.14& 44.31
\\
JoinCL            & 52.57 & 31.42  & 55.03& 29.58 & 35.04&43.85& 41.25
\\
Branch-BERT          & 49.17  & 37.14   & 37.97    & 27.73  & 43.07  &42.13& 39.54
\\
GLAN& \underline{56.95}& \underline{52.38}& \underline{55.98}& 38.15 & \underline{48.91}& \underline{50.42}&\underline{50.47}\\
\hdashline
Llama 2-70b& 49.88   & 46.46    & 43.15     & 39.17& 36.18   &40.23&42.51
\\
ChatGPT(gpt-3.5)& 46.89& 51.69& 53.16  & 36.05 & 27.47 &43.56& 43.14\\
ChatGPT(gpt-4)&49.39&50.71&55.34&\underline{45.09}&40.33 &48.24&48.18
\\
DEEM(gpt-3.5)& 47.97& 48.34& 54.13& 32.46& 42.64& 50.13&45.95
\\
\hline
 \textbf{LLM-CRAN}& \textbf{59.28}& \textbf{54.67}& \textbf{60.32}& \textbf{47.39}& \textbf{49.84}& \textbf{53.62}&\textbf{54.19}
\\
    \hline
    \end{tabular}
    
\end{center}	
\captionsetup{font=small}

\end{table*}

\section{Experimental Setup}
In this section, we describe the evaluation metrics used in the experiments and outline the baseline methods employed for the evaluations.

\subsection{Evaluation Metrics}
We adopt $F_{avg}$ as the evaluation metric to evaluate the performance of stance detection methods, aligning with the approaches in~\cite{li2021p} and~\cite{10.1145/3003433}.
Formally, we first calculate the F1-score for the labels ``\textit{favor}'' and ``\textit{against}'':
\begin{equation}
F_{favor} = \frac{2 \cdot P_{favor} \cdot R_{favor}}{P_{favor} + R_{favor}} 
\end{equation}
\begin{equation}
F_{against} = \frac{2 \cdot P_{against} \cdot R_{against}}{P_{against} + R_{against}} 
\end{equation}
where $P$ and $R$ are precision and recall, respectively. Then, the $F_{avg}$ score is calculated as:
\begin{equation}
F_{avg} = \frac{F_{favor} + F_{against}}{2} 
\end{equation}
We compute the $F_{avg}$ for each target. 

\subsection{Implementation Details}
In our LLM-GRAN model, we use ChatGPT\footnote{https://platform.openai.com/docs/models/gpt-3-5} as the LLM and employ the pre-trained\footnote{https://huggingface.co/google-bert/bert-base-uncased}. The local knowledge layer has a kernel size of 3. In the Multi-hop Attention module, the value of $\lambda$ is set to 0.1, and the number of hops is set to 3. The model is optimized using the Adam optimization algorithm with a batch size of 32 and a learning rate of 0.00001.

\subsection{Baseline Methods}
We perform extensive experiments using state-of-the-art stance detection methods, which fall into four categories: supervised training with DNNs, prompt-tuning with PLMs, fine-tuning with PLMs, and in-context learning with LLMs.

\paragraph{Supervised Training with DNNs}

\begin{enumerate}
\item \textbf{BiLSTM}~\cite{650093} predicts the stance towards a target without explicitly using target information.

\item \textbf{GCAE}~\cite{xue-li-2018-aspect} utilizes a gating mechanism to block target-unrelated information. 

\item \textbf{TAN}~\cite{ijcai2017p557} detects the stances of the posts towards a target by learning the correlation between the posts and the target through the attention mechanism.

\item \textbf{CrossNet}~\cite{xu2018cross} is designed for cross-target stance detection. It encodes the target and the tweet using a BiLSTM and incorporates an aspect attention layer to highlight the core part of a stance-bearing input.
\end{enumerate}





\paragraph{Prompt-tuning with PLMs}

\begin{enumerate}
\item \textbf{MPT}~\cite{KEPrompt} develops prompt-tuning-based PLM for stance detection, with humans defining the verbalizer.

\item \textbf{KPT}~\cite{shin2020autoprompt} introduces external lexicons to define the verbalizer, improving stance detection accuracy.

\item \textbf{KEPrompt}~\cite{KEPrompt} proposes an automatic verbalizer to define label words and a background knowledge injection method to integrate external background knowledge.
\end{enumerate}




\paragraph{Fine-tuning with PLMs}

\begin{enumerate}
\item \textbf{BERT}~\cite{devlin-etal-2019-bert} is a pre-trained language model that predicts stance by appending a linear classification layer to the hidden representation of the ``[CLS]'' token.

\item \textbf{JoinCL}~\cite{liang2022jointcl} employs stance contrastive learning and target-aware graph contrastive learning to generalize target-based stance features to unseen targets.

\item \textbf{TTS}~\cite{li2023tts} utilizes target-based data augmentation to extract informative targets from each training sample and then uses the augmented targets for zero-shot stance detection.

\item \textbf{Branch-BERT}~\cite{li2022improved} learns contextual information within a branch of the conversation for stance detection.

\item \textbf{GLAN}~\cite{niu2024challenge} captures long and short dependencies in conversations using global attention, local convolution, and graph convolution networks. It enhances sentence representations with a multi-hop attention mechanism and incorporates target vectors through a target attention layer, improving stance detection accuracy in multi-turn conversations.
\end{enumerate}






\paragraph{In-context Learning with LLMs}

\begin{enumerate}{}{}
\item  \textbf{Llama}~\cite{devlin-etal-2019-bert}. We conducted experiments using in-context learning with one demonstration sample on a large model. For Llama, we selected Llama 2-70b\footnote{https://huggingface.co/meta-llama/Llama-2-70b-chat-hf} for our experiments.

\item  \textbf{ChatGPT}~\cite{zhang2023investigating}. Similarly, we conducted experiments using in-context learning with one demonstration sample on a large model. For ChatGPT, we selected GPT-3.5\footnote{https://platform.openai.com/docs/models/gpt-3-5} and GPT-4\footnote{https://platform.openai.com/docs/models/gpt-4-and-gpt-4-turbo} for our experiments.

\item  \textbf{DEEM}~\cite{wang-etal-2024-deem-dynamic}. DEEM generates a diverse pool of experts and filters out the experienced ones based on their frequency of appearance and response accuracy. During the inference stage, it dynamically retrieves the experts relevant to the new sentence to determine the stance of the sentence.. 

\end{enumerate}


\begin{table*}
\centering
\fontsize{9pt}{12pt}\selectfont
  \caption{Comparison of different models for cross-target stance detection.}
  \label{tab:cross-target}
\begin{tabular}{lcccccccc}
\hline
     \multirow{2}{*}{\textbf{Method}}&      \multicolumn{8}{c}{\textbf{In-domain}}\\
 \cline{2-9}
\textbf{}&      \multicolumn{2}{c}{DT$\rightarrow$ JB}&\multicolumn{2}{c}{JB$\rightarrow$ DT}&\multicolumn{2}{c}{SX$\rightarrow$ TS}&\multicolumn{2}{c}{TS$\rightarrow$ SX}\\
\hline
     CrossNet&      \multicolumn{2}{c}{14.33}&\multicolumn{2}{c}{15.35}&\multicolumn{2}{c}{20.09}&\multicolumn{2}{c}{17.90}\\
 KEPrompt
& \multicolumn{2}{c}{11.75}&  \multicolumn{2}{c}{13.19}&\multicolumn{2}{c}{20.09}& \multicolumn{2}{c}{31.85}\\
 BERT& \multicolumn{2}{c}{20.34}& \multicolumn{2}{c}{24.87}& \multicolumn{2}{c}{30.06}& \multicolumn{2}{c}{37.32}\\
 TTS& \multicolumn{2}{c}{28.87}& \multicolumn{2}{c}{30.41}&\multicolumn{2}{c}{38.78}& \multicolumn{2}{c}{40.06}\\
GLAN& \multicolumn{2}{c}{\underline{30.10}}& \multicolumn{2}{c}{\underline{31.56}}&\multicolumn{2}{c}{\underline{40.08}}& \multicolumn{2}{c}{\underline{40.85}}\\
 \textbf{LLM-CRAN}& \multicolumn{2}{c}{\textbf{36.28}}& \multicolumn{2}{c}{\textbf{38.84}}&\multicolumn{2}{c}{\textbf{46.24}}& \multicolumn{2}{c}{\textbf{49.12}}\\
\hline
 \multirow{2}{*}{\textbf{Method}}& \multicolumn{8}{c}{\textbf{Cross-domain}}\\
 \cline{2-9}& BC$\rightarrow$ DT & BC$\rightarrow$ JB & BC$\rightarrow$ SX  & BC$\rightarrow$ TS & DT$\rightarrow$ BC & TS$\rightarrow$ BC & SX$\rightarrow$ DT  & DT$\rightarrow$ SX \\
\hline
 CrossNet& 23.47& 21.29& 26.04& 23.73& 23.94 & 21.67 & 12.46 & 11.88 \\
KEPrompt& 30.23& 29.14& \underline{43.72}& 36.68& 15.67 & 24.89 & 23.18 & 23.39 \\
 BERT& 32.45 & 28.34 & 40.26 & 34.84 & 33.21 & 27.65 & 36.08 & 22.89 \\
 TTS& \underline{32.97}& \underline{32.70}& 39.38& 36.37& \underline{35.29}& \underline{37.35}& \underline{39.58}& 26.27 \\
 GLAN& 30.12 & 28.78 & 40.56 & \underline{38.49}& 29.39 & 30.18 & 32.40 & \underline{27.73}\\
 \textbf{LLM-CRAN}& \textbf{34.35}& \textbf{32.92}& \textbf{44.25}& \textbf{41.27}& \textbf{40.81}& \textbf{42.96}& \textbf{40.12}& \textbf{30.51}\\
 \hline
  \end{tabular}
\end{table*}

\begin{table*}
\fontsize{9pt}{12pt}\selectfont
\centering
  \caption{\label{tab:depth} Results of different models at various depths, considering the conversation history for each instance.}
  \begin{tabular}{l|c|ccccclcccc}
    \hline
    \textbf{Target}&\textbf{Depth}& CrossNet            & KEPrompt& BERT &   TTS& Branch-BERT &GLAN
&Llama& ChatGPT& DEEM&\textbf{LLM-CRAN}\\
    \hline
    \multirow{3}{*}{\textbf{Bitcoin}}&1-2& 45.98 & 54.57 & 52.14 &  57.08& 56.85  &56.46 
&50.41 & 48.87 & \underline{58.86}& \textbf{59.65}\\
      &3-5& 36.23  & 55.31 & 54.79 &  51.92 & 49.23   &\underline{59.76}&52.40& 47.38& 47.87&\textbf{60.32}\\
      &6-8& 35.33 & 38.84 & 51.74 &  51.63 & 49.50&\underline{53.99}&44.59 & 38.95 & 38.70&\textbf{58.97}\\
      \hline
      \multirow{3}{*}{\textbf{Tesla}}&1-2
& \underline{41.21}& 28.57 & 33.33 &  \textbf{51.32}& 23.81  &24.44
&35.28& 27.75& 28.57&30.95\\
      &3-5
& 38.56  & 43.43 & 47.24  &  45.79  & 30.76   &49.42  
&46.56& 50.60& \underline{51.03}&\textbf{58.27}\\
      &6-8& 31.80& 46.79& 49.02&  40.94& 40.86 &54.92
&47.12 & \underline{53.76}& 44.29&\textbf{59.00}\\
      \hline
 \multirow{3}{*}{\textbf{SpaceX}}& 1-2
& 35.23 & 49.41 & 47.22 &  50.38 & 41.67  &50.77 
&42.11& 41.65& \underline{57.79}&\textbf{71.35}\\
 & 3-5
& 35.87  & 55.67  & 50.72  &  54.16  & 39.14  &\underline{56.95}&51.03& 54.30& 53.38&\textbf{59.70}\\
 & 6-8& 25.96& 35.53& 37.85&  48.96& 33.28 &53.23
&46.60& 54.30& \underline{54.72}&\textbf{55.07}\\
 \hline
 \multirow{3}{*}{\textbf{Biden}}& 1-2
& 22.86 & 34.21 & 28.49 &  31.43 & 24.14  &28.95 
&\underline{39.57}& \textbf{46.28}& 33.33&20.53\\
 & 3-5
& 23.18  & 30.40& 31.68  &  31.54  & 31.15   &36.46  
&\underline{38.61}& 35.85& 32.27&\textbf{44.67}\\
 & 6-8& 18.65& 29.86& 28.05&  29.94& 24.04 &\underline{42.01}&40.22& 33.61& 30.68&\textbf{49.59}\\
 \hline
 \multirow{3}{*}{\textbf{Trump}}& 1-2
& 24.53 & 31.82 & 32.20&  31.82 & 31.67  &33.33 
&35.49& 24.73& \underline{36.50}&\textbf{37.09}\\
 & 3-5
& 37.83  & 41.50& 47.56 &  \underline{48.14}& 41.97   &50.25  
&37.17& 26.71&  39.09&\textbf{49.35}\\
 & 6-8& 37.87  & 39.58 & 42.77 &  45.05  & 40.76 &\underline{47.35}&34.57 & 26.40& 43.56&\textbf{52.81}\\
 \hline
 \multirow{3}{*}{\textbf{Post-T}}& 2& 35.42& 50.46& 50.29&  43.23&  31.58&\underline{60.31}&54.49& 52.84& 58.88&\textbf{62.06}\\
 & 3-4& 44.27& 47.21& \underline{48.18}&  35.58&  42.52&44.36&43.39& 40.07& 47.84&\textbf{49.47}\\
      &5-6& 33.95& 46.74& 32.21&  40.31&  40.11&\underline{50.03}&20.45& 40.78& 47.50&\textbf{54.16}\\
    \hline 
  \end{tabular}
\end{table*}

\section{Experimental Results}
In this section, we conduct comprehensive experiments on our MT$^2$-CSD dataset. Specifically, we present model comparisons in both in-target and cross-target setups. The reported results are the averages from three different initial runs.

\subsection{In-Target Stance Detection}
We first report the experimental results on the MT$^2$-CSD dataset under the in-target setting. 

Two distinct settings are used in the experiments: considering individual posts or comments as input, and considering the entire conversation history. The results of these experiments are illustrated in Table~\ref{tab:result}.
The results show that our LLM-CRAN outperforms nearly all baseline models on the MT$^2$-CSD dataset. Significance tests comparing LLM-CRAN to Branch-BERT, JoinCL, and TTS indicate that LLM-CRAN achieves statistically significant improvements across most evaluation metrics (with a p-value of $<$ 0.05).

Specifically, analysis of the experimental results yields the following key observations:
First, models that utilize conversational input consistently outperform those using individual sentences as input. Within the same model, incorporating context leads to a performance improvement of 3.55\%. This finding underscores the significance of developing the CSD task.

Second, fine-tuning and prompt-tuning approaches demonstrate substantial improvements over traditional DNNs, highlighting the efficacy of PLMs. 
It is noteworthy that LLM-based methods exhibit suboptimal performance across all targets.
For example, Llama achieves 42.51\%, GPT-3.5 Turbo 43.14\%, GPT-4 48.18\%, and DEEM 45.95\%. This phenomenon can be attributed to inherent limitations of LLMs whose knowledge bases are typically constructed from historical data and may not accurately capture emerging targets or events.
Moreover, LLMs exhibit notable biases in political assessments. For instance, GPT-3.5 scores merely 27.47\% on ``Trump''-related tasks, while DEEM achieves only 32.46\% on ``Biden''-related tasks. 
These discrepancies suggest that LLMs' judgments on political events may be influenced by inherent biases, hindering their ability to comprehend and evaluate based solely on semantic content.

The proposed LLM-CRAN outperforms almost all baseline methods. 
The significance tests comparing LLM-CRAN to Branch-BERT, JoinCL, and TTS reveal that LLM-CRAN exhibits a statistically significant improvement across most evaluation metrics (with a p-value of $<$ 0.05). 
Sepcifically, the proposed LLM-CRAN model demonstrates significant performance improvements over existing approaches. It outperforms the leading DNN-based model (GCAE) by 13.62\%, the most effective prompt-tuning with PLM model (KPT) by 8.94\%, and our previous GLAN model by 3.72\%. 
LLM-CRAN's superior performance can be attributed to its effective utilization of logical and conversational knowledge generated by LLMs. 
This approach enables a more comprehensive understanding of conversations by establishing strong connections between sentences within the conversation context.


\begin{figure}
\centering
\includegraphics[width=0.9\linewidth]{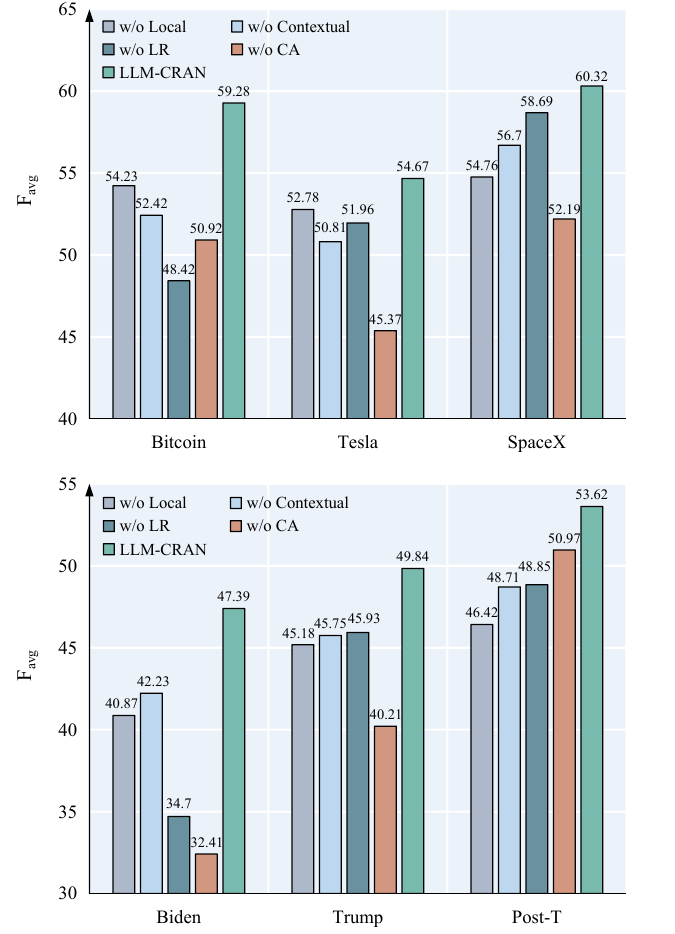}
\caption{\label{fig:Ablation study} Ablation test results.}
\end{figure}

\begin{figure}
\centering
\includegraphics[width=0.9\linewidth]{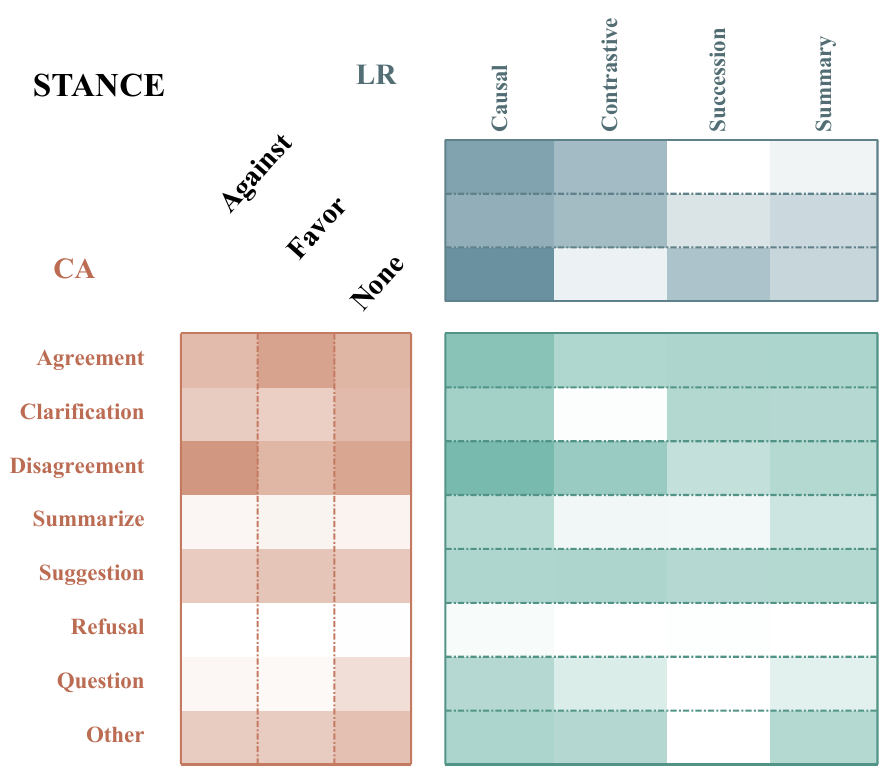}
\caption{Heat maps of the conditional distributions between Logical Relations ({\color{LR}{\textbf{LR}}}), Conversational Acts ({\color{CA}{\textbf{CA}}}), and Stance (\textbf{STANCE}). The {\color{LR}{\textbf{blue}}}, {\color{CA}{\textbf{coral}}} and {\color{STANCE}{\textbf{green}}} maps represent the distributions of {\color{LR}{\textbf{LR}}} conditioned on \textbf{STANCE}, {\color{CA}{\textbf{CA}}} conditioned on \textbf{STANCE}, and {\color{CA}{\textbf{CA}}} conditioned on {\color{LR}{\textbf{LR}}}, respectively.} 
\label{fig:heatmaps}
\end{figure}

\subsection{Cross-Target Stance Detection}
To evaluate the generalizability of our LLM-CRAN model, we conduct cross-target experiments on the MT$^2$-CSD dataset. 
In cross-target experiments, the model is trained on a source target and evaluated on a destination target. Considering the domain differences between the source and destination targets, we further categorized the experiments into two subtasks: in-domain cross-target and cross-domain cross-target. 
We constructed four in-domain cross-target tasks (DT$\to$JB, JB$\to$DT, SX$\to$TS, TS$\to$SX), and 8 cross-domain cross-target tasks 
(BC$\rightarrow$ DT, BC$\rightarrow$ JB, BC$\rightarrow$ SX, BC$\rightarrow$ TS, DT$\rightarrow$ BC, TS$\rightarrow$ BC, SX$\rightarrow$ DT, DT$\rightarrow$ SX). 
In this context, the source target is indicated on the left side of the arrow, while the right side represents the destination target.
The result shown in Table~\ref{tab:cross-target}.

Specifically, for in-domain cross-target setting, LLM-CRAN consistently achieved the highest scores across all experiments, indicating its robustness and ability to leverage contextual and relational information effectively.  
For instance, when trained on JB and tested on DT, it scores 38.84\%, outperforming the best baseline method (GLAN) by 7.28\%. 
Among other models, TTS and GLAN also perform well but generally fall short of LLM-CRAN. This suggests that while TTS and GLAN can be effective in in-domain cross-target settings, LLM-CRAN's integration of logical relations and conversation acts provides a competitive edge.

In cross-domain scenarios, LLM-CRAN consistently surpasses the comparative methods. For instance, when trained on DT and tested on BC, it achieves a score of 40.81\%, outperforming the best baseline method by 5.52\%. This performance enhancement can be attributed to LLM-CRAN's ability to effectively incorporate various external knowledge, which is domain-agnostic.

\begin{table*}
\fontsize{9pt}{12pt}\selectfont
\caption{\label{tab:case} Case study.}
    \begin{tabularx}{\textwidth}{|>{\raggedright\arraybackslash}m{0.1\linewidth}lllllll}
    \hline
    \textbf{Example} & \multicolumn{7}{>{\raggedright\arraybackslash}p{0.85\linewidth}|}{
    \textbf{Post}: SpaceX giant rocket explodes minutes after launch from Texas. \newline
    \textbf{C1}: Understand that when it blew up, the people that built it cheered loudly. Because everything after this thing lighting engines and clearing the pad was a bonus. They now have test data. This was one of many possible “successful” outcomes.\newline
    \textbf{C2}: How it is successful? They wanted a trip around the globe. It landed in the Gulf of Mexico. It was launched from Texas.\newline
    \textbf{C3}: They wanted a trip around the globe. The main goal was to learn, which is what testing is for. That goal was achieved. That is the success.\newline
    \textbf{C4}: Marketing is one hell of a thing, ain’t it? \newline
    \textbf{C5}: Not understanding how an iterative testing process works is a hell of a thing ain’t it?\newline
    \textbf{C6}: Just keep lowering the bar and you’ll eventually be able to call having a rocket at all a success. Don’t confuse learning experience with “success”. A success is what happens when you reach your final goal.} \\
    \hline
    \textbf{True Stance} & Post: Against&C1: Favor &C2: Against &C3: Favor&C4: Against&C5: Favor&C6: Against \\
    \hline
    \textbf{Method}& \multicolumn{7}{c}{\textbf{Predicted Stance}}\\
    \hline
    CrossNet & Post: Against\color{green}{\ding{52}}&C1: Against\color{red}{\ding{56}} &C2: Against\color{green}{\ding{52}} &C3: Against\color{red}{\ding{56}} &C4: Favor\color{red}{\ding{56}}: &C5: Favor\color{green}{\ding{52}} &C6: None\color{red}{\ding{56}}  \\
    \hline
    KEPrompt & Post: Against\color{green}{\ding{52}}&C1: Favor\color{green}{\ding{52}} &C2: Favor\color{red}{\ding{56}}&C3: Favor\color{green}{\ding{52}}&C4: Against\color{green}{\ding{52}}&C5: Against\color{red}{\ding{56}}&C6: Favor\color{red}{\ding{56}}  \\
    \hline
    Branch-BERT & Post: None\color{red}{\ding{56}} &C1: Favor\color{green}{\ding{52}}&C2: Against\color{green}{\ding{52}}&C3: Against\color{red}{\ding{56}}&C4: Against\color{green}{\ding{52}}&C5: Favor\color{green}{\ding{52}}&C6: Favor\color{red}{\ding{56}} \\
    \hline
    GLAN & Post: Against\color{green}{\ding{52}}&C1: Favor\color{green}{\ding{52}}&C2: Against\color{green}{\ding{52}}&C3: Favor\color{green}{\ding{52}}&C4: Against\color{green}{\ding{52}}&C5: Against\color{red}{\ding{56}}&C6: Favor\color{red}{\ding{56}}  \\
    \hline
    ChatGPT & Post: Against\color{green}{\ding{52}}&C1: Favor\color{green}{\ding{52}}&C2: Against\color{green}{\ding{52}}&C3: Favor\color{green}{\ding{52}}&C4: Against\color{green}{\ding{52}}&C5:None\color{red}{\ding{56}} &C6: Favor\color{red}{\ding{56}}  \\
    \hline
    LLM-CRAN & Post: Against\color{green}{\ding{52}}&C1: Favor\color{green}{\ding{52}} &C2: Against\color{green}{\ding{52}} &C3: Favor\color{green}{\ding{52}}&C4: Against\color{green}{\ding{52}}&C5: Favor\color{green}{\ding{52}}&C6: Against\color{green}{\ding{52}} \\
    \hline 
    \end{tabularx}
\end{table*}

\subsection{Impact of Conversation Depth}
To investigate the model's performance at various depths, we conducted an analysis across different depth levels.
We set depth as ``1-2'', ``3-5'', and ``6-8''. 
Particularly, for Post-T target, given that depth 1 instances (posts) serve as the target, we set the depth categories to ``2'', ``3-4'', and ``5-6''. 
The results for different conversation depths are reported in Table~\ref{tab:depth}. 
From the results, it is evident that the LLM-CRAN model performs exceptionally well across different conversation depths, particularly in deeper conversations. This indicates that the LLM-CRAN model has a robust capability to understand and handle complex conversational structures. In contrast, the performance of other models significantly declines as the conversation depth increases. 



\subsection{Ablation Study}

To investigate the influence of different components on the performance of LLM-CRAN, we conducted an ablation test. This involves removing specific components, including the Local Knowledge Layer (denoted as w/o Local), the Contextual Representation Layer (denoted as w/o Contextual), the Logical Relations Layer (denoted as w/o LR), and the Conversation Act Layer (denoted as w/o CA). The results of this ablation study for the proposed LLM-CRAN are presented in Fig. \ref{fig:Ablation study}.

From the results, we can observe that all four components have a significant impact on the performance of LLM-CRAN. 
Notably, the removal of the CA layer resulted in the largest decrease in performance, averaging a drop of 8.84\%. 
The visualization analysis presented in Fig. \ref{fig:heatmaps} provides additional support for these findings. It illustrates a pronounced concentration of the stance categories ``against'' and ``favor'' within the conversation acts of ``agreement'' and ``disagreement''. 
This distinct distribution pattern reveals a strong correlation between stance and conversation acts, further emphasizing the significant role of the CA layer in the model's performance. 
The removal of the LR layer also led to a considerable performance decrease, with an average drop of 6.10\%. 
These findings highlight the importance of logical relations and conversation acts in analyzing conversation threads using LLMs.
Moreover, the removal of the Local and Contextual Representation layers also resulted in performance declines, albeit to varying degrees, underscoring that all four components significantly impact the performance of LLM-CRAN.
As anticipated, integrating all factors yields the optimal performance across all experimental datasets.


\subsection{Case Study}

To better illustrate the effectiveness of our method, we analyzed cases where LLM-CRAN accurately predicted the stance, while most comparison methods failed. 
As shown in Table \ref{tab:case}, we provide a complete conversation thread with the target being ``SpaceX''. 
In this thread, each post and comment's stance towards ``SpaceX'' is assessed. 
The results demonstrate that LLM-CRAN accurately identifies stances throughout the entire conversation thread, particularly excelling in more complex, deeper comments where other models fail. 
For example, in Comment C6, only the LLM-CRAN model correctly predicts the stance as ``Against'', while other models like CrossNet, KEPrompt, Branch-BERT, and ChatGPT fail to do so. 
This superior performance may attributed to LLM-CRAN's unique architecture, which effectively captures logical relations and conversational acts between comments, enhancing its ability to understand and interpret complex conversation threads.
By leveraging these sophisticated layers, LLM-CRAN provides a more accurate and nuanced analysis of stances in multi-turn conversations.

\section{Conclusion}
In this paper, we introduce the MT$^2$-CSD dataset and the LLM-CRAN model to advance conversational stance detection. 
First, we present the MT$^2$-CSD dataset, which supports multi-target, multi-turn conversations. 
To our knowledge, MT$^2$-CSD is the largest dataset of its kind, with a significant portion of comments exceeding six turns, whereas other datasets rarely exceed two turns. 
Second, we propose the LLM-CRAN model, which incorporates logical relations and conversational acts derived from LLMs to guide stance predictions. 
Extensive experiments demonstrate that LLM-CRAN achieves state-of-the-art accuracy across in-target and cross-target scenarios. 
The proposed framework effectively integrates contextual and relational knowledge for conversational stance detection, representing a significant step toward practical systems capable of generalizing to complex multi-turn conversations.
In future work, we will explore extending the MT$^2$-CSD dataset to multimodal scenarios to advance the real-world application of stance detection.


\bibliographystyle{IEEEtran}
\bibliography{ref}

\begin{thebibliography}{10}
\providecommand{\url}[1]{#1}
\csname url@samestyle\endcsname
\providecommand{\newblock}{\relax}
\providecommand{\bibinfo}[2]{#2}
\providecommand{\BIBentrySTDinterwordspacing}{\spaceskip=0pt\relax}
\providecommand{\BIBentryALTinterwordstretchfactor}{4}
\providecommand{\BIBentryALTinterwordspacing}{\spaceskip=\fontdimen2\font plus
\BIBentryALTinterwordstretchfactor\fontdimen3\font minus \fontdimen4\font\relax}
\providecommand{\BIBforeignlanguage}[2]{{%
\expandafter\ifx\csname l@#1\endcsname\relax
\typeout{** WARNING: IEEEtran.bst: No hyphenation pattern has been}%
\typeout{** loaded for the language `#1'. Using the pattern for}%
\typeout{** the default language instead.}%
\else
\language=\csname l@#1\endcsname
\fi
#2}}
\providecommand{\BIBdecl}{\relax}
\BIBdecl

\bibitem{10537616}
B.~Zhang, D.~Ding, Z.~Huang, A.~Li, Y.~Li, B.~Zhang, and H.~Huang, ``Knowledge-augmented interpretable network for zero-shot stance detection on social media,'' \emph{IEEE Transactions on Computational Social Systems}, pp. 1--12, 2024.

\bibitem{zhang2023twitter}
B.~Zhang, D.~Ding, G.~Xu, J.~Guo, Z.~Huang, and X.~Huang, ``Twitter stance detection via neural production systems,'' in \emph{ICASSP 2023-2023 IEEE International Conference on Acoustics, Speech and Signal Processing (ICASSP)}.\hskip 1em plus 0.5em minus 0.4em\relax IEEE, 2023, pp. 1--5.

\bibitem{10.1145/3701716.3715307}
F.~Niu, Y.~Yang, X.~Fu, G.~Dai, and B.~Zhang, ``C-mtcsd: A chinese multi-turn conversational stance detection dataset,'' in \emph{Companion Proceedings of the ACM on Web Conference 2025}, ser. WWW '25.\hskip 1em plus 0.5em minus 0.4em\relax Association for Computing Machinery, p. 769–772.

\bibitem{li2021p}
Y.~Li, T.~Sosea, A.~Sawant, A.~J. Nair, D.~Inkpen, and C.~Caragea, ``P-stance: A large dataset for stance detection in political domain,'' in \emph{Findings of the Association for Computational Linguistics: ACL-IJCNLP 2021}, 2021, pp. 2355--2365.

\bibitem{li2023stance}
A.~Li, B.~Liang, J.~Zhao, B.~Zhang, M.~Yang, and R.~Xu, ``Stance detection on social media with background knowledge,'' in \emph{Proceedings of the 2023 Conference on Empirical Methods in Natural Language Processing}, 2023, pp. 15\,703--15\,717.

\bibitem{ding2024cross}
D.~Ding, R.~Chen, L.~Jing, B.~Zhang, X.~Huang, L.~Dong, X.~Zhao, and G.~Song, ``Cross-target stance detection by exploiting target analytical perspectives,'' \emph{arXiv preprint arXiv:2401.01761}, 2024.

\bibitem{li2023tts}
Y.~Li, C.~Zhao, and C.~Caragea, ``Tts: A target-based teacher-student framework for zero-shot stance detection,'' in \emph{Proceedings of the ACM Web Conference 2023}, 2023, pp. 1500--1509.

\bibitem{villa2020stance}
R.~Villa-Cox, S.~Kumar, M.~Babcock, and K.~M. Carley, ``Stance in replies and quotes (srq): A new dataset for learning stance in twitter conversations,'' \emph{arXiv preprint arXiv:2006.00691}, 2020.

\bibitem{li2022improved}
Y.~Li, H.~He, S.~Wang, F.~C. Lau, and Y.~Song, ``Improved target-specific stance detection on social media platforms by delving into conversation threads,'' \emph{IEEE Transactions on Computational Social Systems}, 2023.

\bibitem{li2023contextual}
Y.~Li, D.~Wen, H.~He, J.~Guo, X.~Ning, and F.~C. Lau, ``Contextual target-specific stance detection on twitter: Dataset and method,'' in \emph{2023 IEEE International Conference on Data Mining (ICDM)}.\hskip 1em plus 0.5em minus 0.4em\relax IEEE, 2023, pp. 359--367.

\bibitem{grosz1986attention}
B.~J. Grosz and C.~L. Sidner, ``Attention, intentions, and the structure of discourse,'' \emph{Computational linguistics}, vol.~12, no.~3, pp. 175--204, 1986.

\bibitem{fillmore2001frame}
C.~J. Fillmore and C.~F. Baker, ``Frame semantics for text understanding,'' in \emph{Proceedings of WordNet and Other Lexical Resources Workshop, NAACL}, vol.~6, 2001, pp. 59--64.

\bibitem{goodwin1990conversation}
C.~Goodwin and J.~Heritage, ``Conversation analysis,'' \emph{Annual review of anthropology}, vol.~19, pp. 283--307, 1990.

\bibitem{niu2024challenge}
F.~Niu, M.~Yang, A.~Li, B.~Zhang, X.~Peng, and B.~Zhang, ``A challenge dataset and effective models for conversational stance detection,'' in \emph{Proceedings of the 2024 Joint International Conference on Computational Linguistics, Language Resources and Evaluation (LREC-COLING 2024)}, 2024, pp. 122--132.

\bibitem{MohammadKSZC16}
S.~Mohammad, S.~Kiritchenko, P.~Sobhani, X.~Zhu, and C.~Cherry, ``Semeval-2016 task 6: Detecting stance in tweets,'' in \emph{Proceedings of the 10th International Workshop on Semantic Evaluation, SemEval@NAACL-HLT, San Diego, CA, USA, June 16-17}, 2016, pp. 31--41.

\bibitem{zhang2020enhancing}
B.~Zhang, M.~Yang, X.~Li, Y.~Ye, X.~Xu, and K.~Dai, ``Enhancing cross-target stance detection with transferable semantic-emotion knowledge,'' in \emph{Proceedings of the 58th Annual Meeting of the Association for Computational Linguistics}, 2020, pp. 3188--3197.

\bibitem{conforti2020will}
C.~Conforti, J.~Berndt, M.~T. Pilehvar, C.~Giannitsarou, F.~Toxvaerd, and N.~Collier, ``Will-they-won't-they: A very large dataset for stance detection on twitter,'' \emph{arXiv preprint arXiv:2005.00388}, 2020.

\bibitem{glandt2021stance}
K.~Glandt, S.~Khanal, Y.~Li, D.~Caragea, and C.~Caragea, ``Stance detection in covid-19 tweets,'' in \emph{Proceedings of the 59th Annual Meeting of the Association for Computational Linguistics and the 11th International Joint Conference on Natural Language Processing (Long Papers)}, vol.~1, 2021.

\bibitem{allaway2020zero}
E.~Allaway and K.~Mckeown, ``Zero-shot stance detection: A dataset and model using generalized topic representations,'' in \emph{Proceedings of the 2020 Conference on Empirical Methods in Natural Language Processing (EMNLP)}, 2020, pp. 8913--8931.

\bibitem{hochreiter1997long}
S.~Hochreiter and J.~Schmidhuber, ``Long short-term memory,'' \emph{Neural computation}, vol.~9, no.~8, pp. 1735--1780, 1997.

\bibitem{zarrella-marsh-2016-mitre}
G.~Zarrella and A.~Marsh, ``{MITRE} at {S}em{E}val-2016 task 6: Transfer learning for stance detection,'' in \emph{Proceedings of the 10th International Workshop on Semantic Evaluation ({S}em{E}val-2016)}.\hskip 1em plus 0.5em minus 0.4em\relax Association for Computational Linguistics, Jun. 2016, pp. 458--463.

\bibitem{igarashi-etal-2016-tohoku}
Y.~Igarashi, H.~Komatsu, S.~Kobayashi, N.~Okazaki, and K.~Inui, ``Tohoku at {S}em{E}val-2016 task 6: Feature-based model versus convolutional neural network for stance detection,'' in \emph{Proceedings of the 10th International Workshop on Semantic Evaluation ({S}em{E}val-2016)}.\hskip 1em plus 0.5em minus 0.4em\relax Association for Computational Linguistics, Jun. 2016, pp. 401--407.

\bibitem{zhou2017connecting}
Y.~Zhou, A.~I. Cristea, and L.~Shi, ``Connecting targets to tweets: Semantic attention-based model for target-specific stance detection,'' in \emph{Web Information Systems Engineering--WISE 2017: 18th International Conference, Puschino, Russia, October 7-11, 2017, Proceedings, Part I 18}.\hskip 1em plus 0.5em minus 0.4em\relax Springer, 2017, pp. 18--32.

\bibitem{wei2019topic}
P.~Wei, W.~Mao, and G.~Chen, ``A topic-aware reinforced model for weakly supervised stance detection,'' in \emph{Proceedings of the AAAI Conference on Artificial Intelligence}, vol.~33, 2019, pp. 7249--7256.

\bibitem{dey2018topical}
K.~Dey, R.~Shrivastava, and S.~Kaushik, ``Topical stance detection for twitter: A two-phase lstm model using attention,'' in \emph{European Conference on Information Retrieval}.\hskip 1em plus 0.5em minus 0.4em\relax Springer, 2018, pp. 529--536.

\bibitem{du2017stance}
J.~Du, R.~Xu, Y.~He, and L.~Gui, ``Stance classification with target-specific neural attention networks.''\hskip 1em plus 0.5em minus 0.4em\relax International Joint Conferences on Artificial Intelligence, 2017.

\bibitem{sun2018stance}
Q.~Sun, Z.~Wang, Q.~Zhu, and G.~Zhou, ``Stance detection with hierarchical attention network,'' in \emph{Proceedings of the 27th International Conference on Computational Linguistics}, 2018, pp. 2399--2409.

\bibitem{LiPLSLWYH22}
C.~Li, H.~Peng, J.~Li, L.~Sun, L.~Lyu, L.~Wang, P.~S. Yu, and L.~He, ``Joint stance and rumor detection in hierarchical heterogeneous graph,'' \emph{{IEEE} Trans. Neural Networks Learn. Syst.}, vol.~33, no.~6, pp. 2530--2542, 2022.

\bibitem{ConfortiBPGTC21}
C.~Conforti, J.~Berndt, M.~T. Pilehvar, C.~Giannitsarou, F.~Toxvaerd, and N.~Collier, ``Synthetic examples improve cross-target generalization: {A} study on stance detection on a twitter corpus,'' in \emph{Proceedings of the Eleventh Workshop on Computational Approaches to Subjectivity, Sentiment and Social Media Analysis, WASSA@EACL 2021, Online, April 19, 2021}.\hskip 1em plus 0.5em minus 0.4em\relax Association for Computational Linguistics, 2021, pp. 181--187.

\bibitem{zhang2025logic}
B.~Zhang, J.~Ma, X.~Fu, and G.~Dai, ``Logic augmented multi-decision fusion framework for stance detection on social media,'' \emph{Information Fusion}, p. 103214, 2025.

\bibitem{devlin-etal-2019-bert}
J.~Devlin, M.-W. Chang, K.~Lee, and K.~Toutanova, ``{BERT}: Pre-training of deep bidirectional transformers for language understanding,'' in \emph{Proceedings of the 2019 Conference of the North {A}merican Chapter of the Association for Computational Linguistics: Human Language Technologies, Volume 1 (Long and Short Papers)}, Jun. 2019, pp. 4171--4186.

\bibitem{liang2022jointcl}
B.~Liang, Q.~Zhu, X.~Li, M.~Yang, L.~Gui, Y.~He, and R.~Xu, ``Jointcl: a joint contrastive learning framework for zero-shot stance detection,'' in \emph{Proceedings of the 60th Annual Meeting of the Association for Computational Linguistics (Volume 1: Long Papers)}, vol.~1.\hskip 1em plus 0.5em minus 0.4em\relax Association for Computational Linguistics, 2022, pp. 81--91.

\bibitem{shin2020autoprompt}
T.~Shin, Y.~Razeghi, R.~L. Logan~IV, E.~Wallace, and S.~Singh, ``Autoprompt: Eliciting knowledge from language models with automatically generated prompts,'' \emph{arXiv preprint arXiv:2010.15980}, 2020.

\bibitem{KEPrompt}
H.~Huang, B.~Zhang, Y.~Li, B.~Zhang, Y.~Sun, C.~Luo, and C.~Peng, ``Knowledge-enhanced prompt-tuning for stance detection,'' \emph{ACM Transactions on Asian and Low-Resource Language Information Processing}, vol.~22, no.~6, pp. 1--20, 2023.

\bibitem{zhang2022would}
B.~Zhang, D.~Ding, and L.~Jing, ``How would stance detection techniques evolve after the launch of chatgpt?'' \emph{arXiv preprint arXiv:2212.14548}, 2022.

\bibitem{zhang2023investigating}
B.~Zhang, X.~Fu, D.~Ding, H.~Huang, Y.~Li, and L.~Jing, ``Investigating chain-of-thought with chatgpt for stance detection on social media,'' \emph{arXiv preprint arXiv:2304.03087}, 2023.

\bibitem{cai2023human}
Z.~Cai, B.~Chang, and W.~Han, ``Human-in-the-loop through chain-of-thought,'' \emph{arXiv preprint arXiv:2306.07932}, 2023.

\bibitem{zhang2024surveystancedetectionsocial}
\BIBentryALTinterwordspacing
B.~Zhang, G.~Dai, F.~Niu, N.~Yin, X.~Fan, S.~Wang, X.~Cao, and H.~Huang, ``A survey of stance detection on social media: New directions and perspectives,'' 2024. [Online]. Available: \url{https://arxiv.org/abs/2409.15690}
\BIBentrySTDinterwordspacing

\bibitem{lan2023stance}
X.~Lan, C.~Gao, D.~Jin, and Y.~Li, ``Stance detection with collaborative role-infused llm-based agents,'' \emph{arXiv preprint arXiv:2310.10467}, 2023.

\bibitem{upadhyaya2025interpretable}
A.~Upadhyaya, W.~Nejdl, and M.~Fisichella, ``Interpretable zero-shot stance detection with proactive content intervention,'' \emph{Information Processing \& Management}, vol.~62, no.~6, p. 104223, 2025.

\bibitem{niu2024multimodal}
F.~Niu, Z.~Cheng, X.~Fu, X.~Peng, G.~Dai, Y.~Chen, H.~Huang, and B.~Zhang, ``Multimodal multi-turn conversation stance detection: A challenge dataset and effective model,'' in \emph{Proceedings of the 32nd ACM International Conference on Multimedia}, 2024, pp. 3867--3876.

\bibitem{kappa}
M.~L. McHugh, ``Interrater reliability: the kappa statistic,'' \emph{Biochemia medica}, vol.~22, no.~3, pp. 276--282, 2012.

\bibitem{li2022survey}
J.~Li, M.~Liu, B.~Qin, and T.~Liu, ``A survey of discourse parsing,'' \emph{Frontiers of Computer Science}, vol.~16, no.~5, p. 165329, 2022.

\bibitem{stolcke2000dialogue}
A.~Stolcke, K.~Ries, N.~Coccaro, E.~Shriberg, R.~Bates, D.~Jurafsky, P.~Taylor, R.~Martin, C.~V. Ess-Dykema, and M.~Meteer, ``Dialogue act modeling for automatic tagging and recognition of conversational speech,'' \emph{Computational linguistics}, vol.~26, no.~3, pp. 339--373, 2000.

\bibitem{10.1145/3003433}
S.~M. Mohammad, P.~Sobhani, and S.~Kiritchenko, ``Stance and sentiment in tweets,'' \emph{ACM Trans. Internet Technol.}, vol.~17, no.~3, jun 2017.

\bibitem{650093}
M.~Schuster and K.~Paliwal, ``Bidirectional recurrent neural networks,'' \emph{IEEE Transactions on Signal Processing}, vol.~45, no.~11, pp. 2673--2681, 1997.

\bibitem{xue-li-2018-aspect}
W.~Xue and T.~Li, ``Aspect based sentiment analysis with gated convolutional networks,'' in \emph{Proceedings of the 56th Annual Meeting of the Association for Computational Linguistics (Volume 1: Long Papers)}.\hskip 1em plus 0.5em minus 0.4em\relax Association for Computational Linguistics, Jul. 2018, pp. 2514--2523.

\bibitem{ijcai2017p557}
J.~Du, R.~Xu, Y.~He, and L.~Gui, ``Stance classification with target-specific neural attention,'' in \emph{Proceedings of the Twenty-Sixth International Joint Conference on Artificial Intelligence, {IJCAI-17}}, 2017, pp. 3988--3994.

\bibitem{xu2018cross}
C.~Xu, C.~Paris, S.~Nepal, and R.~Sparks, ``Cross-target stance classification with self-attention networks,'' in \emph{Proceedings of the 56th Annual Meeting of the Association for Computational Linguistics (Volume 2: Short Papers)}, 2018, pp. 778--783.

\bibitem{wang-etal-2024-deem-dynamic}
X.~Wang, Y.~Wang, S.~Cheng, P.~Li, and Y.~Liu, ``{DEEM}: Dynamic experienced expert modeling for stance detection,'' in \emph{Proceedings of the 2024 Joint International Conference on Computational Linguistics, Language Resources and Evaluation (LREC-COLING 2024)}.\hskip 1em plus 0.5em minus 0.4em\relax ELRA and ICCL, May 2024, pp. 4530--4541.

\end{thebibliography}


\end{document}